\documentclass[runningheads]{llncs}
\usepackage{graphicx}
\usepackage{lipsum}
\usepackage{float}
\usepackage{booktabs}
\usepackage{amsmath}
\begin{document}
\title{Deep Learning for Short-term Instant Energy Consumption Forecasting in the Manufacturing Sector}
\titlerunning{Deep Learning for Short-term Instant Energy Consumption Forecasting}
%
\author{Nuno Oliveira\orcidID{0000-0002-5030-7751} \and
Norberto Sousa\orcidID{0000-0003-2919-4817} \and
Isabel Praça\orcidID{0000-0002-2519-9859}}
\authorrunning{N. Oliveira and N. Sousa et al.}
%
\institute{Research Group on Intelligent Engineering and Computing for Advanced Innovation and Development (GECAD), Porto School of Engineering (ISEP), 4200-072 Porto, Portugal \\ 
\email{\{nunal,norbe,icp\}@isep.ipp.pt}}
\maketitle              
\begin{abstract}

Electricity is a volatile power source that requires great planning and resource management for both short and long term. More specifically, in the short-term, accurate instant energy consumption forecasting contributes greatly to improve the efficiency of buildings, opening new avenues for the adoption of renewable energy. In that regard, data-driven approaches, namely the ones based on machine learning, are begin to be preferred over more traditional ones since they provide not only more simplified ways of deployment but also state of the art results. In that sense, this work applies and compares the performance of several deep learning algorithms, LSTM, CNN, mixed CNN-LSTM and TCN, in a real testbed within the manufacturing sector. The experimental results suggest that the TCN is the most reliable method for predicting instant energy consumption in the short-term.


\end{abstract}
\section{Introduction}

Electricity powers our current world, from illumination and home appliances to cars and industrial machines. Energy consumption is tied to the most varied factors, such as time of year and weather, and due to fundamental issues in storing that much power \cite{yang2013new}, energy production needs to ramp up and down in order to deal with times of high usage while not wasting resources in the downtimes.

This planning and resource management needs to happen, both in the short-term where predictions are made for minutes, hours, weeks and months, and in the long-term where years and decades are provisioned for. Due to the differences involved in both these time ranges and the strategies used in each one, the energy forecasting domain can be divided in two subtopics: short-term and long-term. In the short-term, the focus is on predicting energy consumption values, to aid in planning for energy production and grid maintenance. Whereas in the long-term, planning for grid expansions and future investments are the objective, even influencing the economic development of a given geographic zone \cite{khuntia2018long}.

Although both types of energy forecasting have high importance and impact in everyone's life, the scale of long-term forecasting surpasses most commercial and industrial entities. On the other hand, short-term forecasting can be utilized to improve efficiency of buildings \cite{ZHOU2020114416} and plan around cheaper renewable energy utilization \cite{SALKUTI2019956}, information usable in small scales in the most varied industries.

In short-term energy forecasting, several probabilistic \cite{mori2005probabilistic} and regression \cite{papalexopoulos1990regression} models have been deployed for many years as state of the art approaches to this problem. Now, with the advent of Machine Learning (ML), new techniques are being developed, taking advantage of advances in computing power and Artificial Intelligence (AI) research to improve the old methods and develop new ones. 

Older, more involved physical models, or white-box models, require detailed building information, such as heating, mass and energy consumption, as well as a high amount of manual tuning. When in the initial stages of design or development of a given building, such parameters might not be available. Despite the high setup effort, models of this kind offer a high degree of explainability for their predicted values. Contrarily, data-driven models, or black-box, where ML models are trained using historic consumption data and basic building information to compute future consumption \cite{CHEN20222656}, offer simplicity and straightforwardness in deployment in conjunction with state of the art results, with the caveat that the relation between the input data and output results is not mappable.

In the context of Cyberfactory\#1 \cite{cyberfactory}, an holistic approach to a Factory of the Future, originated energy consumption data from a real life example from the manufacturing industry, namely fabric manufacturing. Therefore, this work proposes a comparison of 4 Deep Learning (DL) models, in a data-driven approach to the energy forecasting domain, applied to a current and relevant dataset.

This work is organized into multiple sections that can be described as follows. Section 2 provides an overview on current state of the art of the energy forecasting domain. Section 3 describes the main characteristics of the used data, as well as of the explored models. Section 4 presents the obtained results performs a comparison between the different models. Finally, Section 5 provides a summary of the main conclusions to be drawn from this work, appointing further research lines.


\section{Related Work}

The employment of ML is not a new endeavor in historical energy forecasting, with several works exploring interesting approaches to this domain, starting with simple model implementations, such as Random Forest (RF) in \cite{7110975} to more sophisticated multi-model ensembles \cite{PINTO2021747}.

However, neural networks dominate the literature, with several different implementations being proposed, such as in \cite{10.1016/j.aei.2016.03.003}, where Che-Jung Chang \textit{et al.} demonstrated a backpropagation neural network, achieving promising results for country-wide eletricity consumption and \cite{luo2019development} where X.J. Luo \textit{et al.} developed a feedforward neural network for load prediction in the context of temperature regulation in buildings. 

A very popular approach sees the use of Long-Short Term Memory (LSTM) model architectures with Convolutional Neural Network (CNN) layers \cite{CNN_LSTM,kCNN_LSTM,gaCNN_LSTM,arimaCNN_LSTM,CNN_LSTM2,CNN_LSTM3}. LSTMs are able to model temporal data and generate predictions from time series, and CNNs are able to extract interesting features from time series predictions. For this reason, this ML pipeline is very utilized in the literature with some of the works improving on it with interesting results. 

In \cite{gaCNN_LSTM}, X.J. Luo \textit{et al.}, employed a genetic algorithm to optimize the architecture of a LSTM, managing to improve over other hyperparameter tuning approaches for the same model in the same dataset.

In \cite{kCNN_LSTM}, Nivethitha Somu \textit{et al.}, introduces a cluster analysis step to the usual CNN-LSTM pipeline, using k-means to understand energy consumption trend by clustering time series data into consistent individual groups. 

In \cite{arimaCNN_LSTM}, Ashutosh Kumar Dubey \textit{et al.}, compared 3 models in forecasting time series data, LSTM, AutoRegressive Integrated Moving Average (ARIMA) and its seasonality aware implementation, Seasonal ARIMA (SARIMA). The authors demonstrate empirically the superiority of the LSTM implementation for the chosen dataset.

Temporal Convolutional Networks (TCN) also produce interesting results due to the time series based nature of the used data. Examples of this can be seen in \cite{9210509}, where a TCN is used in conjunction with a Light Gradient Boosting Machine (LightGBM) to forecast short-term load for several industries in the context of different countries. Furthermore, in \cite{8896684}, H. Wang \textit{et al.}, deploys a TCN on a hourly consumption dataset to great success, surpassing even the reliable LSTM model.

In long-term load forecasting, the involvement of several external socioeconomic factors impacts the predictions made by the usual models. This, allied to the great importance of the results to future developments and investments in a given geographical area, requires more sophisticated approaches.

In \cite{1460110}, T.Q.D. Khoa \textit{et al.}, factors such as gross state product, consumers price index and electricity tariff are used as input multiple neural network based models. The historical data used encompasses 20 years of readings, achieving good results with the Wavelet Network model.
On the other hand, in \cite{nalcaci2019long} Gamze Nalcaci \textit{et al.}, expands the normal approach for short-term forecasting, stretching the predictions for years in the future. The data used contains readings of historical energy consumption from the Turkish electricity distribution network, in conjunction with weather data, and allows not only long-term but also short-term forecasting. Three models were employed, Multivariate Adaptive Regression Splines (MARS), Artificial Neural Network (ANN) and Linear Regression (LR), with MARS achieving more stable results.


\section{Materials and Methods}

In order to determine and compare the short-term energy forecasting capabilities of ML algorithms, four types of ANNs were selected, applied and compared using a real testbed under the manufacturing context. Hence, in this section the main characteristics of the dataset which served as case study are presented along with the main properties of the selected algorithms. Additionally, the employed test design is also discussed in detail, addressing how hyperparameter tuning was performed, how different algorithms were compared and which evaluation metrics were considered. The objectives of this study can be described as follows:

\begin{itemize}
    \item Compare the short-term energy forecasting performance of four ANNs: LSTM; CNN; mixed CNN-LSTM network; and TCN;
    \item Understand the effect of different context windows (an hour, a day and a week) in both algorithm's predictive performance and computational cost;
    \item Conclude which technique is best suited to be used in the context of the Cyberfactory\#1 research project, a fully integrated, real-world, energy forecasting setting.
\end{itemize}

\subsection{Dataset}

In the context of the Cyberfactory\#1 research project, data from 10 energy consumption analysers, expressed in Watts and scattered along a given fabric factory's shop floor, was collected for a period of approximately two years, from the 1st of January, 2020, to the 1st of December, 2021. Energy consumption data was recorded every 4:30 minutes, with small variations in seconds, totalling about 2,200,000 records (220,000 records per each analyser). Table \ref{tab:dataset} presents the monitoring resource of each energy analyser:

\begin{table}[H]
\caption{Energy consumption analysers.}
\centering
\footnotesize
\begin{tabular}{ll}
\toprule
\textbf{Analyser   }	& \textbf{Resource} \\
\midrule
$1^{st}$        & General QGBT              \\
$2^{nd}$        & Compressor                \\
$3^{rd}$        & Jacquard Loom	            \\
$4^{th}$        & Cad Design                \\
$5^{th}$        & Screen/offset painting    \\
$6^{th}$        & Rapier Loom               \\
$7^{th}$        & Cutting Division          \\
$8^{th}$        & Dyeing Machine 1          \\
$9^{th}$        & Dyeing Machine 2          \\
$10^{th}$       & Dyeing Machine 3          \\
\bottomrule
\end{tabular}
\label{tab:dataset}
\end{table}

For this study only data from analysers 6 and 7 was considered in order to reduce significantly the amount of required computational resources to train and optimize the hyperparameters of the selected models. These analysers were selected due to their relevance in the domain, as the cutting division and rapier loom can be seen as core for the factory operation.

\subsection{Cleaning}

Before splitting the dataset and applying further processing operations, an initially cleaning step was performed in order to guarantee data consistency. This step was necessary since the raw measures collected from the energy analysers over time were susceptible to arbitrary errors in sensor readings and communication failures, resulting in anomalous instant energy consumption records such as unrealistically large values or negative ones. Furthermore, communication failures/system failures led to periods of time were no data was captured, resulting in gaps much larger than the expected average of 4:30 minutes between readings. This lack of equally-spaced measures can be troublesome for algorithm's predictive performance when framing the problem as an univariate time series.

In order to clean the dataset and assure data consistency, three operations were applied to the whole dataset:
\begin{enumerate}
    \item \textbf{Cutoff Filter:} This operation was firstly applied to every record in order to remove crystal clear outliers from the dataset. The instant energy consumption measures can not assume negative values, and similarly, values above a certain threshold are so unrealistically large that can only be the product of an erroneous sensor reading. Therefore, using the \textit{Cuttoff Filter} across the whole series, readings bellow a minimum threshold, $\alpha$, were replaced by the value of $\alpha$, and values above a maximum threshold, $\beta$, were replaced by the value of $\beta$. For the employed filter, $\alpha$ was set to 0 and $\beta$ to 10000.
    \item \textbf{Window-based Outlier Substitution:} This filter was applied to replace less obvious outliers and produce a cleaner signal overall. It uses z-score outlier detection, Equation 1, computed over a rolling window, $w$, of $n$ lag measures in order to prevent look-ahead bias i.e, using future measures to decide if the current one should be considered an outlier or not.
    \begin{equation}
        z_{x} = \frac{x-\mu_{w_{x}}}{\sigma_{w_{x}}}
    \end{equation}
    where $z_{x}$ is the z-score of $x$, $\mu_{w_{x}}$ is the mean of all measures of the window $w$ of $n$ lagged values prior to $x$ and $\sigma_{w_{x}}$ is the standard deviation of that same $w$ window. For a given value of $x$, if its z-score, $z_{x}$ is above a given threshold, $\omega$, $x$ is replaced by $\mu_{w_{x}}$. This operation is performed for every element, starting at the $n + 1$ position of the series. For the employed filter, a rolling window of 7 days was applied and the threshold, $\omega$, was set to 3.
    \item \textbf{Data Aggregation:} To deal with the problem of equally-spaced measures, time-based data aggregation was performed in the series as an additional cleaning step. As the raw data was collected at an average rate of one measure every 4:30 minutes, the data aggregation must regard a time window, $t$, at least equal to that same value. Hence, for an initial date, $d_{i}$, to an end date ,$d_{e}$, every instant energy consumption measure between the current date, $d_{c}$, and the next date, $d_{n} = d_{c} + t$, was summed, equalling a new value of aggregated energy consumption measures, $v$, that is indexed along with $d_{n}$ as an entry of a resulting dataset. Then, $d_{c}$ takes the value of $d_{n}$ and the aforementioned operation is repeated until $d_{c}$ becomes greater or equal than $d_{e}$. For the employed operation, a time window, $t$, of 5 minutes was selected. Additionally, the chosen dates for $d_{i}$ and $d_{e}$ were 2020-01-01 00:10H and 2021-12-01 00:00H, respectively.
\end{enumerate}

\subsection{Test design}

After applying every cleaning operation, the resultant dataset was split into three different sets, training, validation and testing, with respect to series continuity. The training set, comprised of 14 months, between 1st of January 2020 and 1st of March 2021, was used to train the four DL algorithms and to  provide a baseline for fitting preprocessing operations such as Min-Max Normalization. Differently, the validation set, comprised of 3 months, between 1st of March 2021 and 1st of June 2021, was used as a testbed for hyperparameter tuning. And, finally, the test set, corresponding to the last 6 months of the dataset, between 1st of June 2021 and 1st of December 2021, was used to evaluate the predictive performance of the already tuned algorithms.

These experiments were repeated for different context windows, one day, one hour and one week, corresponding to different versions of the dataset, with the main purpose of understanding how the increase/decrease of lagged measures affects both the predictive performance of algorithms and the involved computational costs. Consequently, considering two analysers (6 and 7), four DL algorithms (LSTM, CNN, CNN-LSTM and TCN) and three context windows (hour, day, week), the training/validation/testing process was repeated 24 times with an NVIDIA P4000 GPU with 8 Gigabytes of VRAM serving as hardware support.

\subsection{Modeling}

According to the review of the literature, four DL algorithms were selected to be compared in the task of short-term instant energy consumption forecasting. These can be briefly described as follows:

\begin{itemize}
    \item \textbf{LSTM}: An LSTM is a type of Recurrent Neural Network (RNN). Hence, it comprises feedback connections that allow information to travel in a loop from layer to layer, storing information about past computations through a hidden state (network memory). LSTM cells were also designed to overcome the vanishing and exploding gradient problem of typical RNN architectures, making them suitable to learn long-term relationships between elements of the input sequence \cite{app11041674}.
    \item \textbf{CNN}: CNNs are very popular algorithms in image processing and object recognition tasks. These were also adopted for time-series analysis due to their receptive field capability, which allows them to explore efficiently the time dependencies across several events of a given time frame \cite{9699339}.
    \item \textbf{CNN-LSTM}: In a CNN-LSTM network, the receptive field capability of CNNs is used to transform multiple inputs into high-level features that are then processed by consequential LSTM layers. The CNN-LSTM network has shown promising results in recent studies on energy consumption forecasting, in particularly, for problems that regard multivariate time series \cite{KIM201972}.
    \item \textbf{TCN}: The term TCN was first introduced to describe a generic architecture for convolutional sequence prediction. It combines dilations and residual connections with the causal convolutions needed for autoregressive prediction, outperforming generic RNN architectures for a variety of tasks \cite{Bai2018AnEE}.
\end{itemize}

The input of each algorithm is three-dimensional with the shape of ($n_{samples}$, $n_{timesteps}$, $n_{features}$), where $n_{samples}$ is the amount of energy consumption measures (14 months of 5 minute spaced measures for the training set), $n_{timesteps}$ is the number of the considered time window (12 for one hour, 288 for one day, and 2016 for one week) and where $n_{features}$ is the amount of features for each time step. In this case there is only one feature at each time step, the amount of instant energy consumption, as the problem was framed as an univariate time series forecasting task.
    
For each algorithm, several hyperparameters, such as the amount of the hidden layers, learning rate and batch size were optimized using 10 iterations of random search over a predefined grid of parameter values comprising about 500 possible combinations. For the LSTM, the employed hyperparameter grid is presented in Table \ref{tab:lstmhyperparameters}.

\begin{table}[H]
\caption{Hyperparameter grid for the LSTM.}
\centering
\footnotesize
\begin{tabular}{ll}
\toprule
\textbf{Hyperparameter   }	& \textbf{Values} \\
\midrule
Dropout Rate                & [0.1, 0.2, 0.3]       \\
Nº of LSTM Layers       & [2, 3, 4]	            \\
LSTM Layer Size             & [64, 128, 256]        \\
MLP Layer Size              & [32, 64]              \\
Learning Rate               & [1e-4, 1e-3, 1e-2]    \\
Batch Size                  & [32, 64, 256]         \\
\bottomrule
\end{tabular}
\label{tab:lstmhyperparameters}
\end{table}

\subsection{Evaluation}
With respect to the literature, Mean Absolute Error (MAE) and Root Mean Squared Error (RMSE) were selected as evaluation metrics. MAE provides clear interpretation since it provides a direct overview of the average error, while RMSE has the benefit of penalizing large errors since these are first squared before being averaged. Both metrics are expressed as the same unit of measurement as the target value, in this case, Watts \cite{BLAGA2019119,MAWSON2020109966}.

Both MAE and RMSE are presented in Equation 2.

\begin{subequations}
\begin{equation}
    MAE = \frac{1}{n} \sum_{j=1}^{n} |y_j - \hat{y}_j|
\end{equation}    
\begin{equation}
    RMSE = \sqrt{\frac{1}{n} \sum_{j=1}^{n} (y_j - \hat{y}_j)^2}
\end{equation}
\end{subequations}
where $n$ is the total number of measures in the test set, $j$ is a given test set measure, $y_j$ is the actual value of measure $j$ and $\hat{y}_j$ is the predicted value for measure $j$.

\section{Results and Discussion}

The results presented in this section derive from the implementation of the described methodology using Python programming language and it's appropriate libraries: numpy; pandas; scikit-learn; tensorflow and keras.

Each selected algorithm was trained and tested using different context windows, optimizing the appropriate hyperparameters for each scenario using the validation set. Hence, the training duration, in minutes, presented in both Tables \ref{tab:analyser6} and \ref{tab:analyser7}, accounts for the whole process, that is, training, hyperparameter tuning, inference and metric computation.

\begin{table}[H]
\caption{Forecasting results for analyser 6 (duration is presented in minutes).}
\centering
\footnotesize
\begin{tabular}{lllll}
\toprule
\textbf{Algorithm} & \textbf{Window} & \textbf{MAE} & \textbf{RMSE} & \textbf{Duration} \\
\midrule
LSTM        & 12    & \textbf{128.61}   & 353.10            & 59.03     \\
LSTM        & 288   & 129.13            & 347.55            & 750.25    \\
LSTM        & 2016  & 129.69            & 340.89            & 4872.32   \\
CNN         & 12    & 133.46            & 347.34            & 18.41     \\
CNN         & 288   & 146.23            & 350.46            & 20.53     \\
CNN         & 2016  & 140.71            & 354.53            & 64.48     \\
CNN-LSTM    & 12    & 141.45            & 355.16            & 37.52     \\
CNN-LSTM    & 288   & 146.30            & 352.16            & 47.42     \\
CNN-LSTM    & 2016  & 138.54            & 343.41            & 253.98    \\
TCN         & 12    & 134.14            & 349.59            & 42.99     \\
TCN         & 288   & 129.80            & 347.21            & 312.73    \\
TCN         & 2016  & 130.57            & \textbf{340.30}   & 3066.73   \\
\bottomrule
\end{tabular}
\label{tab:analyser6}
\end{table}

For analyser 6 (Table \ref{tab:analyser6}) and regarding the MAE, the LSTM algorithm outperformed every other method for all window sizes, achieving the lowest value, 128.61, for a window of 12 timesteps (one hour). Nevertheless, the RMSE achieved by that same configuration is one of the highest, evidencing the occurrence of large forecasting errors. In that sense, the TCN algorithm achieved the best value of RMSE, 340.30, for a window of 2016 steps (one week), while reaching a MAE of 130.57, slightly higher than the one presented by the LSTM.

\begin{table}[H]
\caption{Forecasting results for analyser 7 (duration is presented in minutes).}
\centering
\footnotesize
\begin{tabular}{lllll}
\toprule
\textbf{Algorithm} & \textbf{Window} & \textbf{MAE} & \textbf{RMSE} & \textbf{Duration} \\
\midrule
LSTM        & 12    & 43.98             & 116.38            & 74.65     \\
LSTM        & 288   & 45.2              & 115.86            & 722.07    \\
LSTM        & 2016  & 46.02             & 117.57            & 4216.95   \\
CNN         & 12    & 44.34             & 115.73            & 16.03     \\
CNN         & 288   & 45.75             & 114.11            & 26.67     \\
CNN         & 2016  & 49.64             & 114.90            & 55.77     \\
CNN-LSTM    & 12    & 43.80             & 115.23            & 36.17     \\
CNN-LSTM    & 288   & 45.35             & 114.85            & 51.78     \\
CNN-LSTM    & 2016  & 48.23             & 115.09            & 155.08    \\
TCN         & 12    & 42.75             & 114.06            & 59.99     \\
TCN         & 288   & \textbf{40.86}    & 110.28            & 533.58    \\
TCN         & 2016  & 43.30             & \textbf{110.14}   & 2523.57   \\
\bottomrule
\end{tabular}
\label{tab:analyser7}
\end{table}

On the other hand, for analyser 7 (Table \ref{tab:analyser7}), the best performing model in terms of both MAE and RMSE, was the TCN, achieving the lowest value of MAE, 40.86, for a window of 288 timesteps (one day) and the lowest value of RMSE, 110.14, for a window of 2016 timesteps. The second and third best values for both metrics were also achieved by the TCN algorithm, making it the clear winner in the context of analyzer 7.

In terms of the context window, there is no evident correlation between the window size and the values of MAE and RMSE. Nonetheless, it is apparent that computational costs substantially increase for all algorithms as the number of timesteps increases as well, being the processing time of both LSTM and TCN considerably bigger than those presented by the remaining approaches. Finally, it can also be concluded that computational-intensive algorithms seem to contribute to better short-term instant energy consumption forecasting capabilities, with the TCN, the second more expensive algorithm, being the best performer overrall for both analysers 6 and 7.

\section{Conclusion}

In this work, the short-term predictive capabilities of four algorithms, a LSTM, a CNN, a CNN-LSTM and a TCN, were evaluated and compared in a real testbed in the manufacturing sector. It was possible to conclude that the TCN was the best overall performer regarding both MAE and RMSE for the two energy analysers considered in the study, which monitored the factory's rapier loom and cutting division for approximately two years. It was also possible to conclude that the computational-intensive algorithms, LSTM and TCN, achieved substantial performance gains when compared with both the CNN and the CNN-LSTM at the price of increased compute duration.

According to the experimental results, there is no apparent correlation between the size of the context windows and the values of MAE and RMSE, so, further energy analysers from the testbed should be included in future works in order to avoid selection bias and reach a more accurate conclusion. 

\section*{Acknowledgements}
The present work has been developed under the EUREKA ITEA3 Project CyberFactory\#1 (ITEA-17032) and Project CyberFactory\#1PT (ANI|P2020 40124) co-funded by Portugal 2020. Additionally, this work has received funding from FEDER Funds through COMPETE program and from National Funds through FCT under the project SPET–PTDC/EEIEEE/029165/2017. Furthermore, this work has also received funding from the project UIDB/00760/2020.


\bibliographystyle{ieeetr}
\bibliography{bibliography}
\end{document}